\documentclass{article}

\PassOptionsToPackage{numbers, compress}{natbib}

\usepackage[final]{mlncp_2023}

\usepackage[utf8]{inputenc} 
\usepackage[T1]{fontenc}    
\usepackage{hyperref}       
\usepackage{url}            
\usepackage{booktabs}       
\usepackage{amsfonts}       
\usepackage{nicefrac}       
\usepackage{microtype}      
\usepackage{xcolor}         
\usepackage{wrapfig}

\usepackage{bm}
\usepackage{enumitem}
\usepackage{algorithm}
\usepackage{algpseudocode}

\usepackage{amsmath}
\usepackage{amssymb}
\usepackage{mathtools}
\usepackage{amsthm}

\DeclareMathAlphabet \mathbfcal{OMS}{cmsy}{b}{n}

 \usepackage{tikz}

\newcommand{\ten}[1]{\mathbfcal{#1}} 
\newcommand{\mat}[1]{\mathbf{#1}}

\title{Real-Time FJ/MAC PDE Solvers via Tensorized, Back-Propagation-Free Optical PINN Training}

\author{%
    Yequan Zhao\textsuperscript{\rm 1, \rm *},
    Xian Xiao\textsuperscript{\rm 2, \rm *}, 
    Xinling Yu\textsuperscript{\rm 1},
    Ziyue Liu\textsuperscript{\rm 1},
    Zhixiong Chen\textsuperscript{\rm 1}, \\
    \textbf{Geza Kurczveil\textsuperscript{\rm 2},
    Raymond G. Beausoleil\textsuperscript{\rm 2}, 
    Zheng Zhang\textsuperscript{\rm 1}} \\
  \textsuperscript{\rm 1} University of California, Santa Barbara\\
  \textsuperscript{\rm 2} Hewlett Packard Labs, Hewlett Packard Enterprise\\
  \textsuperscript{\rm *} Equal Contributions
}

\begin{document}

\maketitle

\begin{abstract}
  Solving partial differential equations (PDEs) numerically often requires huge computing time, energy cost, and hardware resources in practical applications. This has limited their applications in many scenarios (e.g., autonomous systems, supersonic flows) that have a limited energy budget and require near real-time response. Leveraging optical computing, this paper develops an on-chip training framework for physics-informed neural networks (PINNs), aiming to solve high-dimensional PDEs with fJ/MAC photonic power consumption and ultra-low latency. Despite the ultra-high speed of optical neural networks, training a PINN on an optical chip is hard due to (1) the large size of photonic devices, and (2) the lack of scalable optical memory devices to store the intermediate results of back-propagation (BP). To enable realistic optical PINN training, this paper presents a scalable method to avoid the BP process. We also employ a tensor-compressed approach to improve the convergence and scalability of our optical PINN training. This training framework is designed with tensorized optical neural networks (TONN) for scalable inference acceleration and MZI phase-domain tuning for \textit{in-situ} optimization. Our simulation results of a 20-dim HJB PDE show that our photonic accelerator can reduce the number of MZIs by a factor of 1.17$\times 10^3$, with only 1.36 J and 1.15 s to solve this equation. This is the first real-size optical PINN training framework that can be applied to solve high-dimensional PDEs. 
\end{abstract}

\section{Introduction}

Partial differential equations (PDEs) are used to describe numerous science and engineering problems. In practical engineering design, solving a PDE via discretization-based numerical methods (e.g., finite difference or finite elment) normally requires a huge amount of computing resources and run-time due to the resulting large-scale algebraic equations. As a result, traditional PDE solvers are often run on a powerful workstation or HPC platform. Recently, physics-informed neural networks (PINN)\cite{lagaris1998artificial,dissanayake1994neural,raissi2019physics} have emerged as a promising meshless approach to solve high-dimensional or parametric PDEs in both forward and inverse problems. 

While PINN can overcome the curse of dimensionality caused by numerical discretizations, training a realistic PINN is still expensive in many cases, limiting their applications in real-time scenarios where repeated and fast training is required. For instance, in safety verification and control of autonomous systems, a Hamiltonian-Jacobi-Issac (HJI) PDE or a Hamiltonian-Jacobi-Bellman (HJB) PDE has to be solved repeatedly as the sensor data and avoidance specification updates. Training such a PINN on a powerful GPU can take over 20 hours~\cite{bansal2021deepreach,onken2021neural}, whereas there are strict requirements on the latency and energy cost of the embedded computing platforms. This prevents the real-time safety-aware decision making for autonomous systems. In medical imaging such as electrical property tomography~\cite{yu2023pifon}, each training can take dozens of hours, and the measured MRI data is private. It is highly desirable to speed up the training on a local edge device.  This motivates the training of PINNs on light edge devices to enable real-time sensing and decision making. 

Optical neural network (ONN) accelerators provide a promising solution for real-time inference and training~\cite{shen2017deep,Feldmann2021,Shastri2021}. However, training PINNs on photonic chips is very challenging due to three constraints. Firstly, photonic multiply-accumulate (MAC) units such as Mach-Zehnder interferometers (MZIs) are much larger ($\sim$10s of microns) than CMOS transistors, resulting in a low integration density. A real-size PINN with $>10^5$ model parameters can easily exceed the available chip size with the square scaling rule where an $N \times N$ optical weight matrix requires $O(N^2)$ MZIs ~\cite{triMZI,rectangleMZI}. Secondly, it is hard to realize on-chip training on photonic chips. 
Several back-propagation(BP)-free methods are proposed to circumvent the "hardware-unfriendly" nature of error feedback in BP \cite{GuDAC, GuAAAI, filipovich2022silicon, buckley2022general, oguz2023forward}. Unfortunately, these methods are also limited by their scalability issue. 
Thirdly, the loss for PINN training includes higher-order derivatives that require multiple BPs to accurately compute. Due to the inefficiency of \textit{in-situ} BP~\cite{hughes2018training, pai2023experimentally}, an alternative numerical method is needed for photonic implementation.


\textbf{Paper Contributions.} 
This paper proposes the first optical training framework that can handle realistic large-size PINNs on the integrated photonic platform. Our major contributions include:
\begin{itemize}[leftmargin=*]
\vspace{-3pt}
\setlength{\itemsep}{2pt}
\setlength{\parskip}{0pt}
\setlength{\parsep}{0pt}
    \item We employ a BP-free approach using only additional inferences to calculate gradient and derivative estimation, enabling training PINN and solving realistic PDEs on a photonic chip.
    \item We utilize a tensor-compressed format to reduce the number of photonic devices and to improve the convergence of the BP-free optical PINN training framework.
    \item We demonstrate numerical simulation of the optical PINN training method to solve a 20-dimensional HJB PDE. Our method is robust to hardware imperfection and achieves competitive performance while reducing 1.17$\times 10^3$ MZI devices and requiring only 1.36 J and 1.15 s to solve this PDE.
\vspace{-3pt}
\end{itemize}
Our approach greatly advances the state-of-the-art, and it can handle optical training of fully connected networks with sizes up to 1024$\times$1024. This work will pave the way for future real-time and fJ/MAC computing for solving complex high-dimensional PDEs. 

\section{Preliminaries}

    \subsection{Optical Neural Networks (ONN) and Tensorized Optical Neural Networks (TONN)}
    We focus on the ONN~\cite{shen2017deep} architecture with singular value decomposition (SVD) to implement matrix-vector multiplication (MVM), i.e., $y=\bm{W}x=\bm{U\Sigma V^*}x$. The unitary matrices $\bm{U}$ and $\bm{V^*}$ are implemented by MZIs in Clements mesh~\cite{rectangleMZI}. The parametrization of $\bm{U}$ and $\bm{V^*}$ is given by $\bm{U}(n)=\bm{D} \prod_{i=2}^n \prod_{j=1}^{i-1} \bm{R}_{i j}\left(\phi_{i j}\right)$ where $\bm{D}$ is a diagonal matrix, and each 2-dimensional rotator $\bm{R}_{ij}(\phi_{ij})$ can be implemented by a $2\times 2$ MZI containing two phase shifters and two 50/50 splitters. We denoted all programmable phases as $\bm{\Phi}$ and $\bm{W}$ is parametrized as $\bm{W}(\bm{\Phi})$. 
    
    
    To increase the scalability of ONN, a tensorized optical neural network (TONN)\cite{xiao2021large} is proposed to realize large-scale ONNs with reduced hardware resources (i.e., MZIs) using the tensor-train (TT) decomposition algorithm.
    Let $\bm{W} \in \mathbb{R}^{M\times N}$ be a generic weight matrix in a neural network. We factorize its dimension sizes as $M = \prod^{L}_{i=1}m_i$ and $N = \prod^{L}_{j=1}n_j$, fold $\bm{W}$ into a $2L$-way tensor $\mathbfcal{W} \in \mathbb{R}^{m_1\times m_2 \times \dots \times m_L \times n_1 \times n_2 \times \dots \times n_L}$, and parameterize $\ten{W}$ with the TT decomposition \cite{oseledets2011tensor}:
        \begin{equation}
        {\mathbfcal{W}}(i_1, i_2, \dots, i_L, j_1, j_2, \dots, j_L)
        \approx \prod^{L} \nolimits_{k=1} \mat{G}_k(i_k, j_k)
        \label{TT decomposition}
        \end{equation}
    Here $\mat{G}_k(i_k, j_k) \in \mathbb{R}^{r_{k-1} \times r_{k}}$ is the $(i_k, j_k)$-th slice of the TT-core $\mathbfcal{G}_k \in \mathbb{R}^{r_{k-1}\times m_k \times n_k \times r_k}$by fixing its $2$nd index as $i_k$ and $3$rd index as $j_k$. The vector $(r_0, r_1, \dots, r_{L})$ is called TT-ranks with the constraint $r_0 = r_{L} = 1$. This TT representation reduces the number of unknown variables from $\prod^{L}_{k=1}m_k n_k$ to $\sum _{k=1}^{L}r_{k-1}m_k n_k r_k$. 
    The detailed architecture of TONN can be found in \cite{xiao2021large}.
    
    \subsection{Physics-Informed Neural Networks (PINNs) and Tensor-compressed PINNs}

    Consider the well-posed initial value partial differential equation (PDE) problem described by:
\begin{equation}
\begin{aligned}
\mathcal{N}[\bm{u}(\bm{x},t)]&=l(\bm{x}, t), &&\bm{x} \in \Omega,~~t \in[0, T],\\
\mathcal{I}[\bm{u}(\bm{x},0)] &= g(\bm{x}), &&\bm{x} \in \Omega,  
\end{aligned}
\label{general PDE}
\end{equation}
where $\bm{x}$ and $t$ are the spatial and temporal coordinates; $\Omega \subset \mathbb{R}^{D}$ and $T$ denote the spatial domain and time horizon, respectively; $\mathcal{N}$ is a general nonlinear differential operator; $\mathcal{I}$ represents the initial condition; $\bm{u} \in \mathbb{R}^{n}$ is the solution for the PDE described above. In PINNs~\cite{raissi2019physics}, a neural network $\bm{u}(\bm{x},t;\bm{\theta})$, parameterized by $\bm{\theta}$, is substituted into PDE \eqref{general PDE}, resulting in a residual defined as:
\begin{equation}
r(\bm{x},t;\bm{\theta}):=\mathcal{N}[\bm{u}(\bm{x},t;\bm{\theta})]-l(\bm{x}, t).
\label{PDE residual}
\end{equation}
The parameters $\bm{\theta}$ can be obtained by minimizing the loss
$
\mathcal{L}(\bm{\theta})=\mathcal{L}_r(\bm{\theta})+\lambda\mathcal{L}_0(\bm{\theta}),
$
where
\begin{equation}
\mathcal{L}_r (\bm{\theta}) = \frac{1}{N_{r}}\sum_{i=1}^{N_{r}}  \left\|r(\bm{x_{r}}^{i},t_{r}^{i};\bm{\theta})\right\|_{2}^{2}~~\text{and}~~ \mathcal{L}_0 (\bm{\theta}) = \frac{1}{N_{0}}\sum_{i=1}^{N_{0}}\left\|\mathcal{I}[\bm{u}(\bm{x_{0}}^{i},0;\bm{\theta})] - g(\bm{x_{0}}^{i})\right\|_{2}^{2},
\label{loss terms}
\end{equation}
are the residuals of the PDE and the initial (or terminal) condition, respectively. To adapt PINNs to the edge with constraints in memory, computation, and energy, \cite{liu2022tt} introduced a tensor-compressed PINN framework, where fully-connected layers are decomposed into a series of TT-cores, as in \eqref{TT decomposition}.


    \begin{figure}[t]
    \centering
    \includegraphics[width=0.74\textwidth]{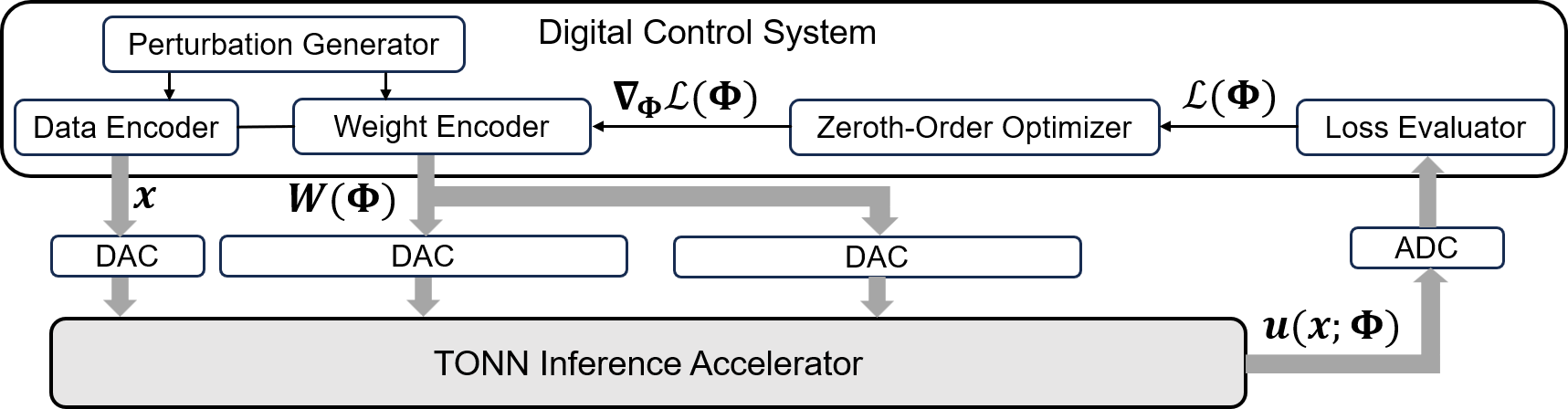}
    \caption{\label{Figure1} The overall architecture of the BP-free optical training accelerator.}
    \end{figure}
\section{BP-Free Optical Training Accelerator for Tensor-Compressed PINN}

    \subsection{Overall Architecture}
         \begin{figure}[h]
    \centering
    \includegraphics[width=0.9\textwidth]{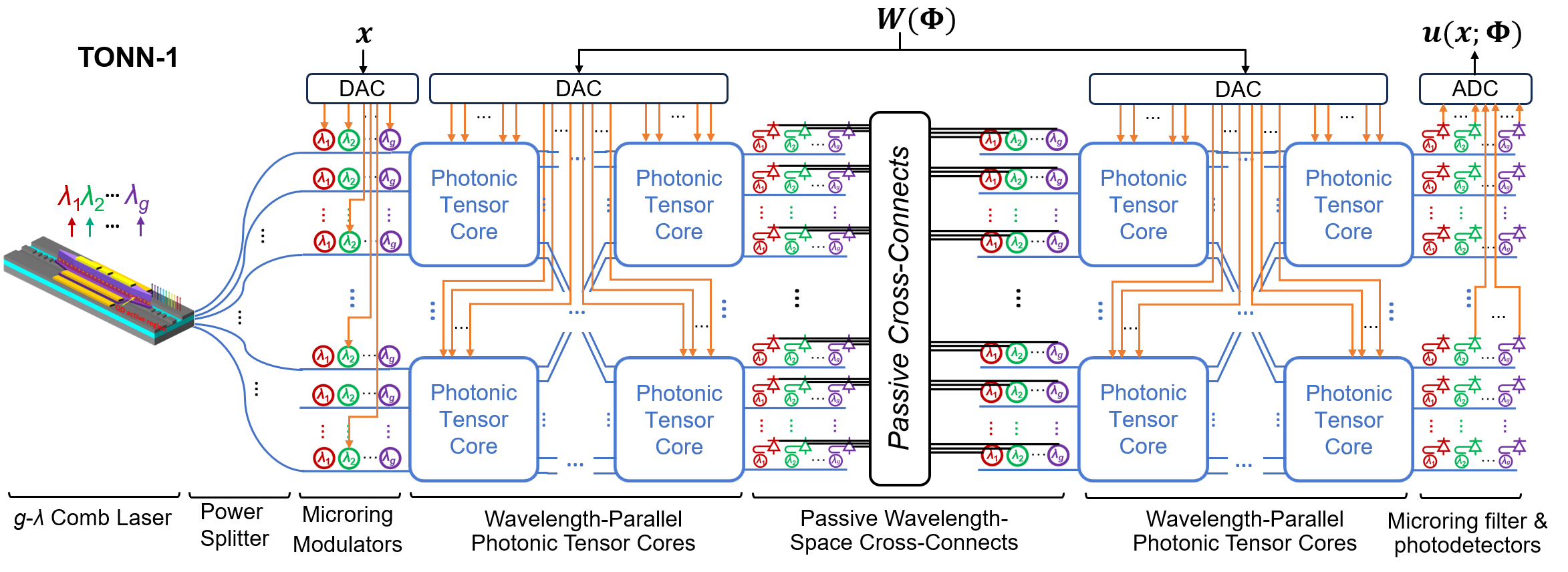}
    \caption{\label{Figure2} TONN-1: The designed tensor-compressed optical inference accelerator based on the TONN architecture with wavelength and space multiplexing.}
    \end{figure}

    The block diagram of our optical PINN training accelerator is shown in Fig.~\ref{Figure1}. This accelerator does not perform any BP. Instead, it repeatedly call an optical inference accelerator TONN\cite{xiao2021large} to obtain some loss information. The collected loss information is process in a digital control system, and gradient information is estimated via a zeroth-order optimization to update PINN model parameters. Since propagations are not used, intermediate results will not be computed or stored on the photonic chip. 
    
    In the following, we give the details of our TONN design and BP-free training method.


    \subsection{Tensor-compressed Optical Inference Accelerator Design}
    We present two designs of optical neural networks based on tensor-train (TT) compression. The TT-based optical neural network design can greatly reduce the number of photonic devices, latency and energy cost. Furthermore, it can reduce the number of on-chip training variables and improve the convergence of the on-chip training framework.

    \begin{wrapfigure}{r}{0.42\textwidth}
    \centering
    \includegraphics[width=0.42\textwidth]{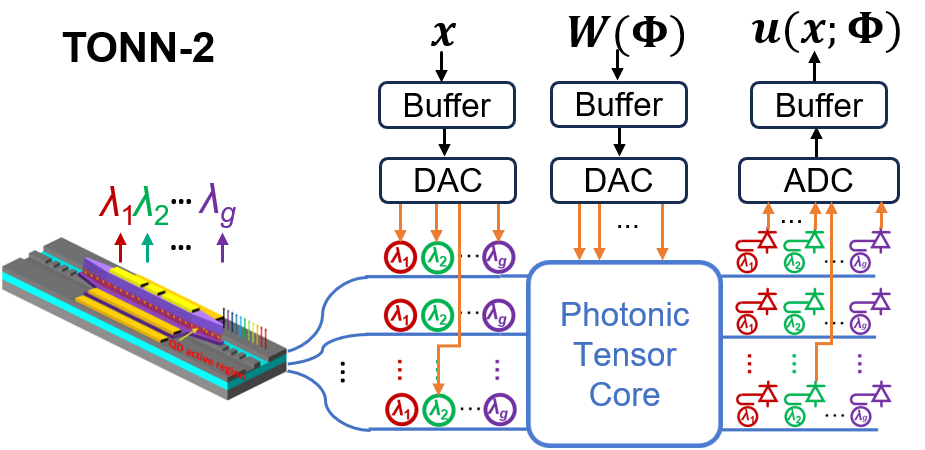}
    \caption{\label{Figure3} TONN-2: The designed inference accelerator using a single wavelength-parallel photonic tensor core with time multiplexing.}
    \vspace{-10pt}
    \end{wrapfigure}
    
   The first design, called TONN-1, is illustrated in Fig.~\ref{Figure2}. In this design, the tensor multiplications between the input data and all tensor-train cores (the whole tensorized matrix) are realized in a single clock cycle by cascading the photonic tensor cores in the space domain and adding parallelism in the wavelength domain\cite{xiao2021large}.

    The second design, called TONN-2 and shown in Fig.~\ref{Figure3}, uses a single wavelength-parallel photonic tensor core~\cite{Xiao2023} with time multiplexing. Compared with TONN-1, TONN-2 exhibits a smaller footprint at the expense of higher latency and additional memory requirements. In each clock cycle, the photonic tensor core with parallel processing in the wavelength domain is updated to multiply with the input tensor. Then, the intermediate output data is stored in the buffer for the next cycle.

    \subsection{Tensor-Compressed BP-Free PINNs Training} \label{Tensor-Compressed BP-Free PINNs Training}
    We adopt the idea of \cite{zhao2023tensor} to implement a fully BP-free PINN training method to mitigate the memory bottleneck of on-chip photonic computing. In PINN training, the BP process should be avoided in both the loss function evaluation and in the SGD-type optimization step.

    \paragraph{BP-free Loss Evaluation.} The differential operator in (\ref{general PDE}) involves first-order and high-order derivatives of $\bm{u}$ with respect to $\bm{x}$. It is hard to compute these derivatives via a BP process on a photonic chip. Two methods can be used to address this issue. The first method is finite difference, which calculates the derivatives by perturbing each element of $\bm{x}$. An alternative method uses sparse-grid Stein estimator~\cite{zhao2023tensor}. Both method only require a few additional inferences with coordinate-wise perturbed input data to estimate first- and second-order derivatives, then compute $\mathcal{L}(\bm{\Phi})$. MZIs do not need to be re-programmed when estimating the derivatives.

   \paragraph{BP-free Gradient Estimation in SGD-type Optimizers.} Stochastic gradient descent (SGD) and its variants are the mainstream optimizer for neural network training. In our optical PINN training framework,
    we use a zeroth-order gradient estimator, Simultaneous Perturbation Stochastic Approximation (SPSA) \cite{spall1992multivariate} to obtain a randomized estimation of the gradient. Specifically,  given a model parameterized by $\bm{\Phi}\in \mathbb{R}^d$ and a loss function $\mathcal{L}$, SPSA computes a randomized gradient estimation

        \begin{equation}
           \hat{\nabla}_{\bm{\Phi}}\mathcal{L}(\bm{\Phi})=
            \sum \nolimits_{i=1}^N \frac{1}{N\mu} \left[\mathcal{L}\left(\bm{\Phi}+\mu \bm{\xi}_i\right)-\mathcal{L}(\bm{\Phi})\right] \bm{\xi}_i.
        \label{ZO gradient estimation}
        \end{equation}

        Here $\{\bm{\xi}_i \in \mathbb{R}^d\}_{i=1}^N$, are $N$ i.i.d. samples drawn from $\mathcal{N}(0, \bm{I}_d)$ and $\mu$ is the sampling radius. In practice, we further adopt the concept from signSGD~\cite{bernstein2018signsgd} and its ZO counterpart, ZO-signSGD~\cite{liu2019signsgd}, to de-noise the SPSA gradient estimation by preserving only the sign for each update. Specifically, given a learning rate $\alpha$, the PINN model parameters are updated as
        \begin{equation}
     \bm{\Phi}_t \leftarrow \bm{\Phi}_{t-1}-\alpha {\rm sign}(\hat{\nabla}_{\bm{\Phi}}\mathcal{L}(\bm{\Phi}))       
        \end{equation}

    Note that we fully leverage the benefits of the tensor-compressed model in both inference and training. The neural network $\bm{u}_{\bm{\Phi}}(\bm{x},t)$ is parameterized by all programmable MZI phases $\bm{\Phi}$ in each photonic TT-core $\ten{G}_k(\bm{\Phi}_k)$. The photonic TT-cores that approximate a weight matrix are directly employed in the inference and updated in the training. Since the gradient variance of the SPSA method grows as the dimensionality of training variables increase, the tensor-compressed format can dramatically reduce the gradient estimation variance and improve the convergence of the ZO training framework.

    SPSA requires $N$ additional loss evaluations to estimate the gradients. During the training process, after evaluating $\mathcal{L}(\bm{\Phi})$, the digital control system generates a perturbation vector and program all MZIs simultaneously. Then, the same training data is shed into the inference accelerator again to conduct the additional inferences $\mathcal{L}(\bm{\Phi + \mu \bm{\xi}}_i)$. After $N$ additional inferences, the digital control system averages over the $N$ loss values and then estimates the gradient $\hat{\nabla}_{\bm{\Phi}}\mathcal{L}(\bm{\Phi})$, finally updates all MZIs with their updated value simultaneously.

    

\section{Experiments Results}
    We evaluate our proposed BP-free tensor-compressed PINN training by training a PINN arising from high-dim optimal control of robots and autonomous systems. We consider the following 20-dim HJB PDE:
       \begin{equation}
        \begin{aligned}
        &\partial_t u(\bm{x}, t)+\Delta u(\bm{x}, t)-0.05 \left\|\nabla_{\bm{x}}u(\bm{x}, t)\right\|_{2}^{2}=-2, \\
        &u(\bm{x}, 1)=\left\|\bm{x}\right\|_{1}, \quad \bm{x} \in [0,1]^{20}, ~~t \in[0, 1].
        \end{aligned}
        \end{equation}
        Here $\left\|\cdot\right\|_{p}$ denotes an $\ell_p$ norm. The exact solution is $u(\bm{x},t)=\left\|\bm{x}\right\|_{1}+1-t$. The baseline neural network is a 3-layer optical neural network ($21\times n, n\times n, n\times 1$, $n$ denotes the number of neurons in the hidden layer) with sine activation. We approximate the solution by a transformed neural network $u(\bm{x},t;\bm{\Phi}) = (1-t)f(\bm{x},t;\bm{\Phi}) + \left\|\bm{x}\right\|_{1} $, where $f(\bm{x},t;\bm{\Phi})$ is the base neural network or its TT-compressed version. We remark that the transformed network is designed to ensure our approximated solution exactly satisfies the terminal condition.
    \subsection{Numerical Simulation Results}

    All numerical simulations are based on a software implementation built upon PyTorch~\cite{paszke2019pytorch} backend and TorchONN library~\cite{gu2021l2ight} to simulate the computational model of an optical computing platform. To show the effectiveness and robustness of our design, we compared our method with different training paradigms. \textbf{Off-chip Training} denotes first pre-training on electrical digital platforms, e.g., CPUs and GPUs, then mapping the trained model to photonic devices. The gradients w.r.t. model parameters and the derivatives w.r.t. the input are computed by BP. Our proposed tensor-compressed BP-free training belongs to \textbf{On-chip Training} as it directly tunes photonic devices (i.e., phase-shifters in MZIs) on-chip and trains from scratch. 
    For off-chip training, we implemented hardware-aware training that incorporates various hardware imperfections and its counterpart hardware-unaware training that runs on an ideal computational model. The hardware-aware training is a hardware-restricted learning problem, where we considered phase-shifter $\gamma$ coefficient drift $\bm{\Gamma}\sim \mathcal{N}(\gamma, \sigma^2_\gamma)$ \cite{GuAAAI, on2021analysis} caused by fabrication variations and thermal cross-talk between adjacent devices $\bm{\Omega}$ \cite{GuAAAI, on2021analysis, zhu2020countering}, and phase bias due to manufacturing error $\bm{\Phi}_b \sim \mathcal{U}(0,2\pi)$ and the objective became $\bm{\Phi}^* = {\arg \min}_{\bm{\Phi}} \mathcal{L}(\bm{W}(\bm{\Omega \Gamma \Phi}+\bm{\Phi}_b))$. For on-chip training, we incorporate the same hardware imperfections to mimic the actual analog hardware.

    \begin{table}[htbp]
      \centering
      \caption{Software simulation results. Both ONN and TONN are three-layer MLPs with sine activation. Off. denotes off-chip training, On. denotes on-chip training, w/o and w/ noise denote hardware-unaware and -aware training, respectively. For off-chip training, we reported the validation loss after mapping to hardware with noise and the original validation loss (in parentheses). For on-chip training, we reported the final validation loss.}
        \begin{tabular}{cccccc}
        \toprule
        \toprule
        Network & Neurons & Params & \multicolumn{1}{c}{Off. w/o noise} & \multicolumn{1}{c}{Off. w/ noise} & On. w/ noise \\
        ~ & ~ & ~ & ~ & ~ & (proposed) \\
        \midrule
        ONN & 1024 & 608,257 & 3.10E-01 (7.63E-03) & 3.07E-01 (7.81E-03) & 1.43E-02 \\
        \midrule
        TONN & 1024 & 1,536  & 3.73E-01 (1.46E-02) & 2.97E-01 (1.35E-02) & \textbf{5.53E-03} \\
        \bottomrule
        \bottomrule
        \end{tabular}%
      \label{tab:loss}
    \end{table}%

    Our results are provided in Table:\ref{tab:loss}. 
    We report the validation loss which is the mean square error (MSE) w.r.t. the ground truth. 
    After training, our proposed BP-free tensor-compressed PINN training achieves a validation loss of 5.53E-3, indicating that the model fits the ground truth well. The tensor-compressed ONN outperforms the un-compressed ONN, indicating that our tensor-compressed training is capable of preserving the expressive power of a wide ONN with greatly reduced model parameters (396$\times$ fewer in this case).

    Off-chip training achieves a similar validation loss on the pre-trained model. However, after mapping to real photonic devices, the performance greatly degrades due to the hardware imperfection. The hardware-aware training does not help significantly as the imperfection model in software is not identical to real hardware. Our proposed method inherently circumvents this problem as it directly tunes on the fabricated hardware during on-chip training, thus demonstrating better robustness and better performance.

    \subsection{System Performance}
    The system performance for the accelerators based on ONNs and TONNs are evaluated and compared, as shown in Table:\ref{tab:sys}, assuming the III-V-on-Si device platform~\cite{9794616}. Since we only use several additional inferences to estimate the gradients and derivatives, a multiplication of number of inference and energy consumption or latency per inference indicates the energy and training efficiency, respectively. For the TONN design, the first two MLP layers are both factorized as 1024*1024=[4*8*4*8]*[8*4*8*4] with TT-ranks as [1,2,1,2,1]. The total number of wavelengths used is 32~\cite{xiao2021large}. The SVD implementation of the arbitrary matrices is considered in the calculation.
    \begin{table}[htbp]
      \centering
      \caption{Comparison of the \# of MZIs, energy/inference, latency, and photonic footprint}
        \begin{tabular}{ccccccc}
        \toprule
        \toprule
        Network & Params & \# of MZIs & Energy & Latency & Footprint \\
         &  &  & /inference (J) & /inference (ns) & (mm$^{2}$) \\
        \midrule
        ONN & 6.08E05 & 2.10E06 & - & 600 & 2.62E05 \\
        \midrule
        TONN-1 & 1.53E03 & 1.79E03 & 6.45E-09 & 550 & 648 \\
        TONN-2 & 1.53E03 & 28 & 5.05E-09 & 3604 & 26 \\
        \bottomrule
        \bottomrule
        \end{tabular}%
      \label{tab:sys}%
    \end{table}%
    
        \textbf{Energy Consumption:} The total energy of the accelerators is mainly consumed in the ADC, DAC, and digital control systems. Here, we focus on the photonic energy consumption per forward, which consists of five parts: laser wall-plug power, microring modulator power, MZI mesh power, microring add-drop filter power, and PD receiver power. The conventional ONN has insurmountable optical loss due to the square scaling rule, so the energy cannot be calculated. The TONN-2 consumes slightly less energy per forward due to lower insertion loss even though it requires 64 cycles.
        
        \textbf{Latency:} The latency per inference in on-chip training is calculated by: $t_{\rm{inference}}=n_{\rm{cycle}}*(t_{\rm{DAC}}+t_{\rm{tuning}}+t_{\rm{opt}}+t_{\rm{ADC}})+t_{\rm{DIG}}$, where $t_{\rm{DAC}}$ is the ADC conversion delay ($\sim$24 ns), $t_{\rm{tuning}}$ is the metal-oxide-semiconductor capacitor (MOSCAP) phase shifter tuning delay ($\sim$0.1 ns), $t_{\rm{opt}}$ is the propagation latency of optical signal ($\sim$51.2 ns for ONN, $\sim$1.6 ns for TONN-1, and $\sim$0.4 ns for TONN-2), $t_{\rm{ADC}}$ is the DAC delay($\sim$24 ns), and $t_{\rm{DIG}}$ is the digital computation overhead ($\sim$500 ns) for gradient calculation and phase updates. The TONN-2 uses 64 cycles for one inference, while ONN and TONN-1 only needs one cycle.
        
        \textbf{Training Efficiency:} In our 20D-HJB example, we need 42 inferences for each loss evaluation and 10 loss evaluations for gradient estimation. Suppose a mini-batch size of 100, 4.20E4 inferences are required for one epoch. The energy consumption per epoch is estimated as 2.71E-04 J and the latency per epoch is estimated as 0.23 ms for TONN-1. On average training reaches a good solution after 5000 epochs, which corresponds to 1.36 J and 1.15 s for solving a 20D-HJB equation.
        
        \textbf{Footprint:} Only the footprint of the photonic devices, which occupy the major area of the accelerator, is used for comparison. The photonic footprint includes the areas of hybrid silicon comb laser, microring resonator (MRR) modulator arrays, photonic tensor cores, MRR add-drop filters, photodiodes, and electrical cross-connects. It can be seen that TONN-2 occupies a much smaller footprint than TONN-1 at the expense of much higher computational latency.

\section{Conclusion}
    In this work, we have proposed the first optical training framework that can handle realistic large-size PINNs on the integrated photonic platform. 
    By introducing a tensor-compressed BP-free training method, we have implemented a large-scale optical inference accelerator with significant hardware and energy reductions and an on-chip training framework that only requires additional inferences for gradient and derivative estimation, leading to scalable and robust optical PINN training. Through numerical simulations on a 20-dimensional Hamiltonian-Jacobi-Bellman (HJB) PDE, our method has shown impressive model size reduction (1.17$\times 10^3$ fewer MZIs), ultra-low-energy (1.36J) and ultra-high-speed (1.15s) PINN training. Future research includes further scaling up our PINN training framework, investigating high-speed MZI tuning methods, and demonstrating an electro-photonic integrated system for fJ/MAC high-speed PDE solvers.

\newpage
\bibliographystyle{unsrt}
\bibliography{mlncp_2023}

\end{document}